\documentclass[letterpaper, 10 pt, conference]{ieeeconf}  % Comment this line out if you need a4paper

\IEEEoverridecommandlockouts                              % This command is only needed if 
\usepackage[labelformat=simple]{subcaption}
\usepackage{todonotes}
\usepackage{graphicx}
\usepackage[hidelinks]{hyperref}

\title{\LARGE \bf
Safety of the Intended Driving Behavior Using Rulebooks
}

\author{Anne Collin$^{1}$, Artur Bilka$^{1}$, Scott Pendleton$^{1}$ and Radboud Duintjer Tebbens$^{1}$% <-this % stops a space
\thanks{*This work was supported by the Hyundai-Aptiv Autonomous Driving Joint Venture.}% <-this % stops a space
\thanks{$^{1}$ Hyundai-Aptiv Autonomous Driving Joint Venture, 100 Northern Avenue, 02210 Boston, MA
        {\tt\small {\{anne.collin, artur.bilka, scott.pendleton, radboud.tebbens\}@aptiv.com}}}%
}

\begin{document}

\maketitle

\thispagestyle{empty}
\pagestyle{empty}

%%%%%%%%%%%%%%%%%%%%%%%%%%%%%%%%%%%%%%%%%%%%%%%%%%%%%%%%%%%%%%%%%%%%%%%%%%%%%%%%
\begin{abstract}

Autonomous Vehicles (AVs) are complex systems that drive in uncertain environments and potentially navigate unforeseeable situations. Safety of these systems requires not only an absence of malfunctions but also high performance of functions in many different scenarios. The ISO/PAS 21448 \cite{InternationalOrganizationforStandardization2019} guidance recommends a process to ensure the Safety of the Intended Functionality (SOTIF) for road vehicles. This process starts with a functional specification that fully describes the intended functionality and further includes the verification and validation that the AV meets this specification. For the path planning function, defining the correct sequence of control actions for each vehicle in all potential driving situations is intractable. In this paper, the authors provide a link between the Rulebooks framework, presented by \cite{Censi2019}, and the SOTIF process. We establish that Rulebooks provide a functional description of the path planning task in an AV and discuss the potential usage of the method for verification and validation.
\end{abstract}

%%%%%%%%%%%%%%%%%%%%%%%%%%%%%%%%%%%%%%%%%%%%%%%%%%%%%%%%%%%%%%%%%%%%%%%%%%%%%%%%

\section{Introduction}
Although research in the design, assessment and validation of fully driverless autonomous vehicles (AVs) is growing, the AV industry recognizes that existing automotive methods are insufficient to demonstrate the performance and safety of self-driving cars \cite{Kalra2016, Fraade-blanar2018}. Specifically, the prevailing standard for automotive functional safety focuses solely on hazards related to electrical and electronic malfunctions \cite{InternationalOrganizationforStandardization2018}, but does not consider hazards due to functional insufficiencies. 
Unlike traditional cars, AVs are responsible for the entire dynamic driving task, including perception and decision making. This responsibility increases the relevance of the safety and performance of functionality.

Recent efforts towards standardization of methods to address potential functional insufficiencies led to the publication of the ISO/PAS 21448 guidance, which defines the concept of Safety of the Intended Functionality (SOTIF) \cite{InternationalOrganizationforStandardization2019}. SOTIF is defined as "the absence of unreasonable risk due to potentially hazardous behaviors related to the intended functionality or performance limitations of a system that is free of faults addressed in the ISO 26262 series" [1, p. iv].
The recommended SOTIF process starts with the functional and system specifications. However, defining the functionality of an AV system remains an open problem for urban driving and many other applications where the complexity exceeds that of dealing with a few well-defined tasks (e.g., a robot moving boxes in a factory on a predefined path without encountering dynamic obstacles). Indeed, the automotive industry largely applies systems engineering processes for defining functionality of individual components or functions; a self-driving vehicle brings another level of complexity, which can have an effect on safety \cite{Sheard2017}. The approach of using a series of conditions and actions, such as "if there is a moderate to severe crash, then the airbag shall deploy", is not compatible with the artificial intelligence that practitioners want to grant to a self-driving vehicle. Conversely, validating a fully machine-learned AV system is challenging because such a system does not contain any explicit specification. Indeed, if a collision were to occur, the behavior of the vehicle would not trace back to a specific definition, which may be unacceptable for society \cite{Amodei2016}.

Furthermore, the quantity of driving situations required to fully cover the Operational Design Domain (ODD, i.e., the domain in which the vehicle is designed to function \cite{Thorn2018}) for an AV's urban driving task is innumerable even in highly restricted operations. If one defines a scenario as a temporal sequence of actions and scenes \cite{Ulbrich2015}, the combinatorial number of possible scenarios makes it practically impossible to specify the desired behavior in each individual scenario. Beyond the work required to specify behavior in a large number of scenarios, this method may overlook situations or particular combinations of specifications in which the AV cannot satisfy the intended behavior or in which the intended behavior remains undefined. 
The lack of common performance metrics beyond the number of human interventions per mile driven hinders the definition of measurable requirements for the system and results in a slow learning feedback loop during the development process.

An additional hurdle in the definition and validation of AV functional specification is the lack of uniform definition of the correct driving behavior in different situations. The existing traffic laws target human drivers and rely on human interpretation. In many countries, the official rules of the road define quantitative constraints only for a few specific situations, such as braking distances \cite{DepartmentofMotorVehicles2018, SGTraffic2017}. Most traffic laws are imprecise and some conflict with each other, leaving it to the best judgment of the driver to ensure safety of all road users. Moreover, different driving styles exist and lead to different behaviors. The Rulebooks framework, presented in \cite{Censi2019}, shows how an order of pre-defined, logical rules for the AV leads to a selection of preferred trajectories. This paper shows the potential use of this formal logic framework for behavior specification. 
%This framework provides a level of abstraction that removes the need for scenario-dependent requirements.

\begin{figure}[h]
	\includegraphics[width=\linewidth]{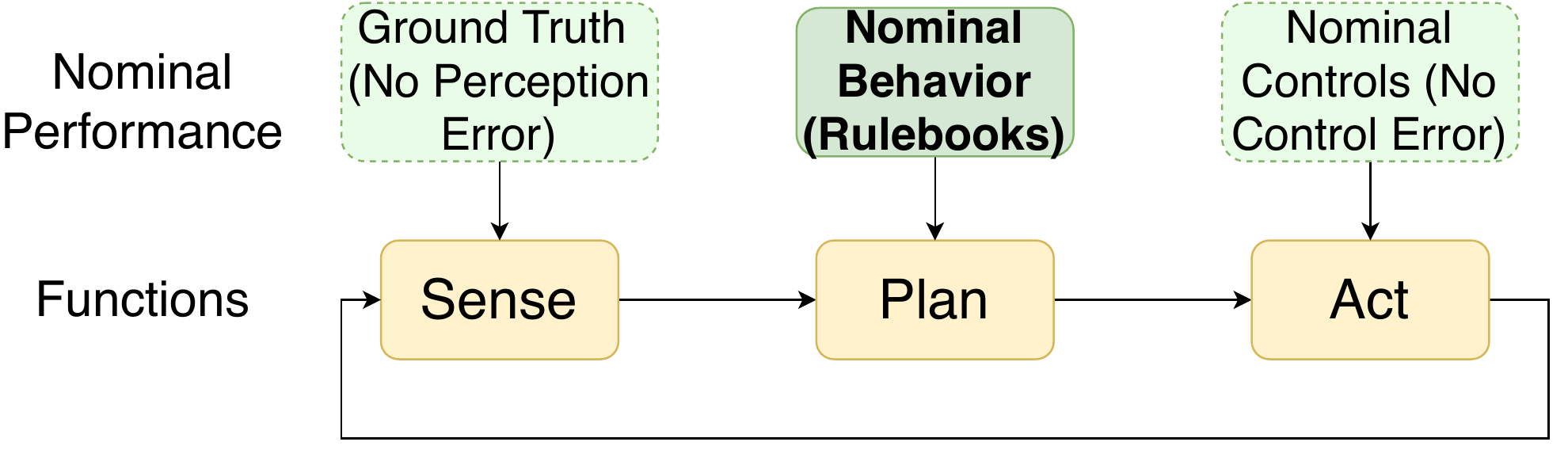}
	\caption{Rulebooks in the context of the Sense-Plan-Act paradigm, in which the robot needs to reevaluate its environment each time it moves. Rulebooks capture the ideal performance of the planning function.}
	\label{fig:SPA}
\end{figure}

The specification traces the AV behavior to the designer's intended functionality. As such, the Rulebooks framework addresses the functional specification, verification, and validation steps of the SOTIF process. This paper focuses on Society for Automotive Engineers (SAE) level 4 AVs, which can operate without human assistance within a defined ODD \cite{SAE2018}. While we recognize the SOTIF challenges related to the perception function for AVs, this work focuses on the AV's path planning function; the main goal is to specify the AV's desired behavior in the context of perfect information (nominal behavior) and to be able to assess to what extent the AV satisfies this desired behavior (Figure \ref{fig:SPA}).

Section \ref{sec:relWork} reviews related literature and section \ref{sec:Rulebook} provides an overview of the Rulebooks framework. Section \ref{sec:RBforSOTIF} demonstrates the use of Rulebooks as part of the SOTIF processes, including functional specification (subsection \ref{subsec:FuncSpec}), verification (subsection \ref{subsec:Ver}) and validation (subsection \ref{subsec:Val}) of the SOTIF. Section \ref{sec:Ex} illustrates the proposed method with a concrete example of behavior specification in a restricted ODD. Section \ref{sec:Concl} summarizes key insights and identifies areas for future work.

\section{Related work and contributions} \label{sec:relWork}
Several frameworks exist related to safety and behavior specification for AVs. The ISO/PAS 21448 guidance provides a process and work products to demonstrate the SOTIF of AV systems, but does not describe how to specify the behavior itself \cite{InternationalOrganizationforStandardization2019}. \cite{APTIV2019} summarizes a complete approach for the safety and validation of SAE level 3 and 4 AV systems but also does not provide details of specific methods for behavior specification. Another still developing effort towards a comprehensive standard for the safety analysis tools and techniques for AVs emphasizes the need for acceptable planning capability, documentation, verification, validation, and risk mitigation but does not offer an approach to achieve this \cite{EdgeCaseResearch2019}. 

The Responsibility-Sensitive Safety (RSS) model \cite{Shalev-Shwartz2017} defines responsibilities and behaviors for road users in specific situations involving collision risks with other agents. The Safety Force Field provides a computational mechanism for obstacle avoidance \cite{Nister2019}. To the authors' knowledge, they are one of the few methods tackling AV behavior definition without fully fixing all variables in a scenario, however they do not provide a means to resolve any conflicts among traffic laws or other requirements. 

Methods exist to keep track of the AV's capabilities during the development process, based on a list of maneuvers defined per hazardous scenario \cite{Nolte2018}. A limitation of this approach is that the resulting capabilities and maneuvers only address the enumerated hazardous scenarios but not new situations the AV may encounter. Singapore Standards Council's Technical Reference 68 acknowledges the "dilemmas" that can occur in legislation regarding conflicting rules and advises the adoption of a minimum violation policy via a rules hierarchy \cite{SG_TR68_2019}. This approach aligns well with our proposal in this work but does not provide context about the relation to safety guidance such as SOTIF.

Numerous studies discuss the challenges for the verification and validation of AVs. These broadly deal with the two complementary types of approaches also identified in ISO/PAS 21448 \cite{InternationalOrganizationforStandardization2019}. First, quantitative approaches use evidence from random and statistically representative exposure to different driving situations on public roads (or possibly in simulation) to make statistical inferences about the safety of the AV in unknown hazardous scenarios. This approach comes with challenges because of (i) the large amount of driving required to obtain high confidence in the AV's rate of rare events like fatal collisions, (ii) the uncertainty about the acceptable level of statistical safety, (iii) the unclear statistical implications of changes in the AV system or its ODD, (iv) the inability to directly observe collisions while testing with a safety driver, and (v) the lack of established proxy metrics to rapidly assess safety (i.e., leading measures) during the iterative development and validation cycle suggested in ISO/PAS 21448 \cite{InternationalOrganizationforStandardization2019, Kalra2016, Junietz2019}.
While quantitative approaches can contribute to safety validation and several studies explore statistical methods to address the challenges with quantitative approaches \cite{Shalev-Shwartz2017, Asljung2017, Fraade-blanar2018, Bin-Nun2020, Zhao2017a}, a safety argument based solely on statistical evidence will likely remain insufficient. 

Second, qualitative, scenario-based testing approaches use controlled testing in simulation or on a closed course to demonstrate high safety performance in known scenarios. Challenges with scenario-based testing include (i) the difficulty of extrapolating the results from staged tests to real-world safety performance, (ii) the absence of consensus on the number of scenarios required, and (iii) the absence of general criteria to assess whether the AV behavior in a scenario is acceptable \cite{APTIV2019}. While research to date offers approaches to generate scenarios and define appropriate completeness metrics, none offer a universal method to define the desired AV behavior in different scenarios \cite{DeGelder2020, Pegasus2019, Foretellix2020, ASAM2019}.  
Consistent with current guidance, \cite{InternationalOrganizationforStandardization2018, APTIV2019}, we believe that addressing the SOTIF requires a complimentary approach that combines statistical and scenario-based testing to address the limitations of each individual method.

A large body of research focuses on defining, and finding in a reasonable amount of time, ideal trajectories for AVs in specific situations. To cite a few, \cite{Parseh2019} proposes a behavior that minimizes collision severity in the event that a collision is unavoidable. \cite{Qian2014} deals with intersections, and \cite{Ulbrich2013} with lane changing. Each would correspond to an implementation solution to enforce a specific rule, however, none of these provide a framework to combine these definitions into behavior-governing rules. In contrast, cost-based planning algorithms like real-time Rapidly-exploring Random Trees (RRT*) can accommodate planning according to explicit rules for driving behavior \cite{Karaman2011, Censi2019, Vasile2017, Tumova2013}. While prior work on rule-based behavior focuses on real-time planning, the method presented in this paper focuses on behavior definition during development, and after-the-fact evaluation.
The Rulebooks, first introduced in \cite{Censi2019}, offers a framework to combine rules. Although the authors formalize mathematically how to deal with the contradictions and interpretability of laws, they do not link this theoretical framework with systems engineering methods recommended by the industry. 

The main contributions of this paper are:

\begin{itemize}
\item The conversion of a theoretical framework into a method to specify an AV's intended driving functionality while following the ISO/PAS 21448 process.
%\item The introduction to the AV safety community of a systematic, transparent, and scenario-independent framework to define the behavior of an AV. This framework can allow AV companies to specify the intended driving functionality.
\item A concrete method to verify that the AV behavior respects safety considerations, traffic laws, comfort, and local driving habits, through an assessment tool based on the Rulebooks framework.  
\item A path for guiding behavior testing towards more difficult scenarios for validation.
\end{itemize}

\section{Rulebooks principles} \label{sec:Rulebook}

This section provides an overview of behavior specification according to the Rulebooks method; we refer to \cite{Censi2019} for further details.

Many factors influence driving behavior. In addition to safety, the specification of AV behavior should take into account traffic laws, the local driving culture, and driving performance (e.g., comfort, time to destination). Adopting a predictable driving style furthermore creates a safer interaction with all the actors sharing the road. In some driving situations, these factors do not fully specify the driving behavior, leaving the choice of trajectory to a company or even a single person's judgment. For other driving situations, respecting all the rules leaves no feasible trajectory, potentially causing the car to stop and disrupt the flow of traffic. The Rulebooks approach considers all factors and addresses the latter situation (i.e., potentially disrupting traffic flow) by leaving some degree of flexibility.

Rulebooks provides a formal method to specify the desired behavior of an AV. A Rulebook contains a set of formal rules derived from traffic laws, locally acceptable driving behaviors, and AV stakeholder needs. Each rule has an associated violation metric on the possible trajectories. The violation metric expresses the degree of violation of the trajectory with respect to the rule. A Rulebook provides a priority structure on the rules. One proposed priority structure is a \textit{pre-order}. Loosely formulated, this hierarchical priority structure specifies the order of importance of rules. In a pre-ordered set, it is possible for two or more rules to remain incomparable, meaning that we do not specify which rule is most important. Alternatively, if there is a clear priority rank preference, the Rulebook can convey this by forcing an order in the Rulebook, providing specificity where desired. A Rulebook expresses an implementation- and scenario-agnostic set of preferences for AV behavior: if it places safety above comfort, then the Rulebook does not care how the AV satisfies this specification, and the preference does not suddenly change because we are in a different scenario.

\cite{Censi2019} proves that a pre-ordered set of rules leads to a pre-order on the set of trajectories. This means that if we define a set of general rules in a Rulebook, then we can rank any set of trajectories into a pre-ordered set that reflects the preferences defined in the Rulebook. Given that the Rulebook specification is scenario-agnostic, we can rank candidate trajectories within any scenario based on the same Rulebook.  This not only eliminates the need to write scenario-specific behavior requirements, but also ensures that a Rulebooks-based AV makes decisions that are internally consistent. Since the Rulebook provides a pre-order on trajectories, some can remain incomparable with each other, leaving flexibility for the AV to select among those trajectories. 

\begin{figure}[h]
	\centering
	\includegraphics[height=9.5cm]{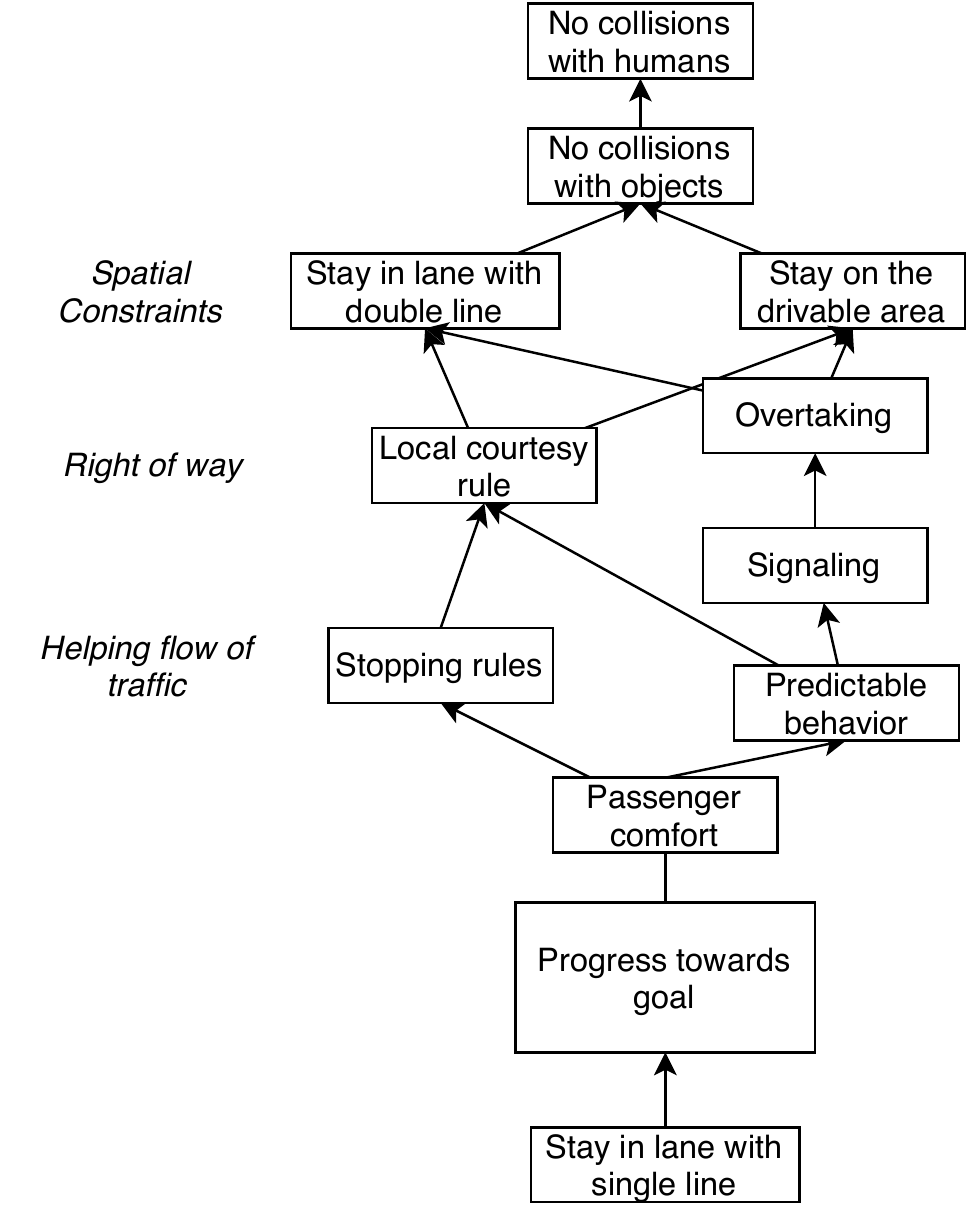}
	\caption{An example of a Rulebook, based on \cite{Censi2019}. It provides a concrete basis for discussing the desired driving behavior.}
	\label{fig:RBExample}
\end{figure}

Figure \ref{fig:RBExample} provides an illustrative example of a Rulebook.

Note that \cite{Censi2019} furthermore provide guidelines on refinement of Rulebooks. Allowable refinements include priority refinement (i.e., the clarification of priorities between previously incomparable rules), rule aggregation (i.e., the collapsing of two or more equally ranked rules), and rule augmentation (i.e., the addition of another rule at the lowest level of priority). These operations could apply to geographic refinements from local vs. federal legislation, or could allow AV developers to specify "house rules" for specialized behaviors beyond traffic laws or regulations.

Note that rules should be third party observable, meaning that they are independent of the internal information state of the AV such as its visibility region. A planning algorithm may need to make some informed prediction about the future state of the world to best mitigate the risk of rule violation, but the violation metric for a given, executed trajectory should come from the observation of actual world states. This leaves the algorithmic implementation flexible; while a developer may choose to inform their methods based on explicit prediction and cost evaluation according to a Rulebook, they could also freely choose to rely heavily on machine learning methods so long as the resultant trajectory performs well in the sense of low Rulebook violation. 

\section{Rulebooks for SOTIF} \label{sec:RBforSOTIF}

\begin{figure}[h]
	\includegraphics[width=\linewidth]{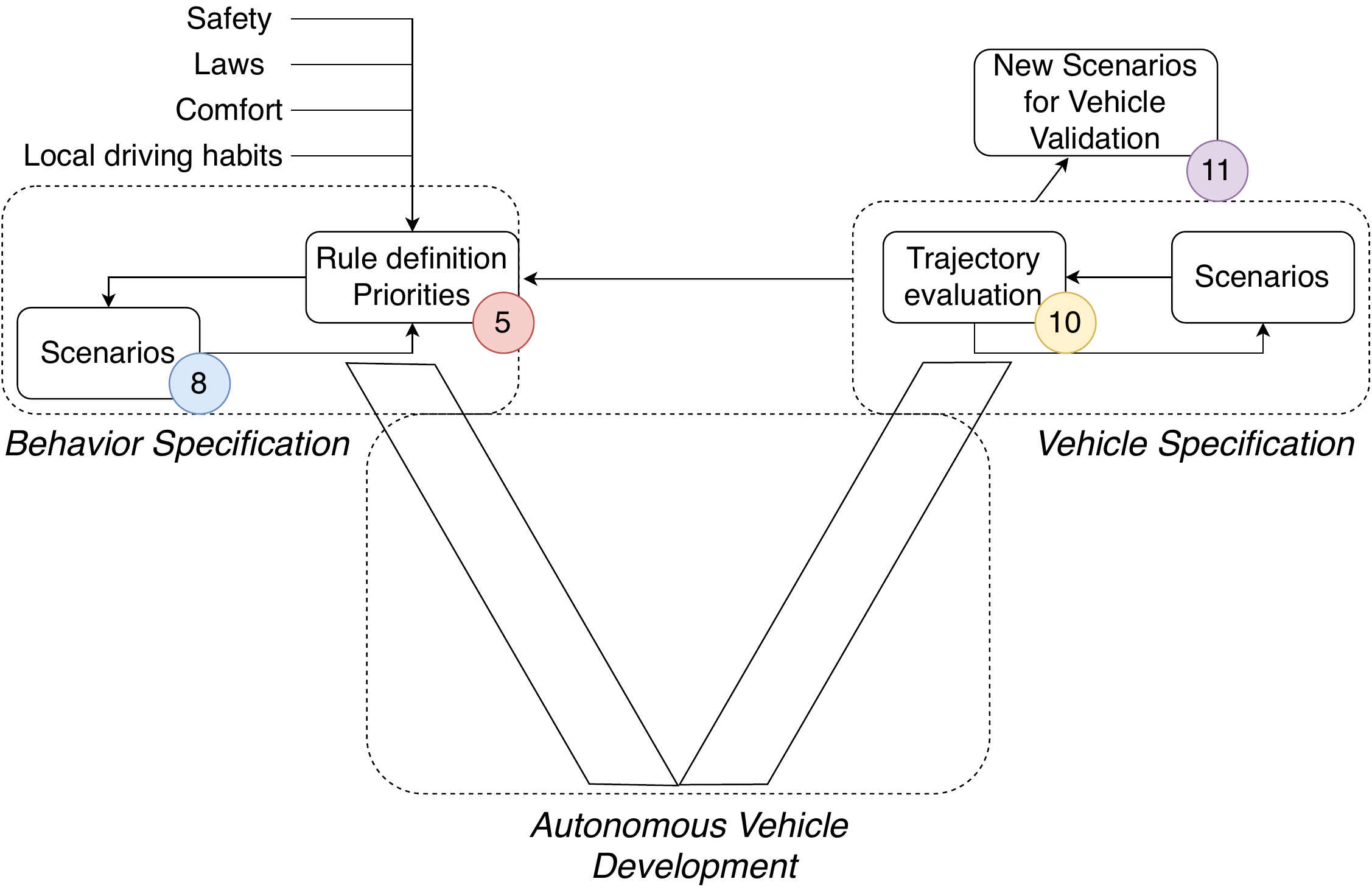}
	\caption{Integration of Rulebooks with ISO/PAS 21448 activities and the V-model in systems engineering. The colored numbers indicate the related clause number in \cite{InternationalOrganizationforStandardization2019}.}
	\label{fig:21448Process}
\end{figure}

This section describes the link between Rulebooks and SOTIF for the steps shown in Figure \ref{fig:21448Process}.

\subsection{Functional Specification} \label{subsec:FuncSpec}

ISO/PAS 21448 recommends that the safety evaluation of hazards related to driving behavior starts with defining the behavior itself. Rulebooks supports this definition in the following ways.

First, Rulebooks offer traceability between its rules and the rules of the road. Like human drivers, AVs should follow local traffic laws and road rules  even if not directly relevant to collision avoidance. In addition to traffic laws, Rulebooks explicitly specify rules regulating comfort, driving style, and other stakeholder needs. As framed in \cite{Stellet2019}, the Rulebooks method aligns the \textit{required} and \textit{specified} behaviors.  
Individual rules encode how the AV should interact with the road infrastructure and other users, and the priority structure provides a full specification of the intended driving functionality in the event that satisfaction of all rules is not possible.  The combination of individual rules and a priority structure defines the preference among possible trajectories. This covers most of the functional description listed in paragraph 5.2 of the ISO/PAS for an AV, namely the goals and description of the intended functionality, and the dependencies and interactions with other road users, environmental conditions, and road infrastructure. Behavior specification for an AV is only valid in a specific ODD \cite{SAE2018}. This is reflected in the rules themselves, which will change as a function of applicable jurisdiction and local driving habits.

Second, Rulebooks can help clarify the architecture requirements by providing a way to score the driving performance of the overall vehicle according to the rules and priorities. One can then derive requirements for subsystems such as localization and mapping \cite{Collin2019}, and perception \cite{Czarnecki2018} by relating the subsystem performance to the overall driving performance.

Third, given that the behavior specification is independent of the actual implementation of the planner, one can verify the specification on its own merits before attempting it on the vehicle. This effectively decouples errors due to poor specification from errors due to poor implementation. Verification of a Rulebook prior to implementation adds a further layer of safety and shortens the development cycle.

The following paragraphs describe how Rulebooks enable the verification and validation of the SOTIF. We use the terms \textit{verification} and \textit{validation} as described in the ISO/PAS 21448 guidance: "verification activities address mainly Area 2 [known unsafe scenarios], whereas validation activities address mainly Area 3 [unknown unsafe scenarios], including the validation of SOTIF in unknown use cases" [1, p.6].
%in the following manner, meant to reflect the ISO/PAS 21448 usage; in this work, \textit{verification} deals with ensuring the system behaves as expected with sufficient coverage in known scenarios, whereas \textit{validation} is the "set of activities gaining confidence that an item is able to accomplish its expected functionalities and mission. \cite{InternationalOrganizationforStandardization2019}.

\subsection{Verification of the SOTIF}\label{subsec:Ver}

The main contribution of Rulebooks towards the verification of the SOTIF is the ability to score a vehicle's trajectories with respect to the behavior specification in known scenarios. Indeed, as shown in \cite{Censi2019}, the priority structure of the rules leads to an ordering of potential trajectories, revealing which one best adheres to the behavior specification. Furthermore, one can specify the acceptable degree of violation of each rule and verify that any violation of rules by the AV is consistent with the Rulebooks' priority structure. 
% Each trajectory can therefore be scored, giving the development teams a goal to reach in their implementation. %We still need to answer how to give a holistic score to a trajectory when the Rulebook is a pre-order, so I think we should omit this for now

The scenarios can come from various sources, including System-Theoretic Process Analysis (STPA) \cite{Leveson2018}, adaptive testing \cite{Corso2019}, orthogonal arrays \cite{hedayat2012orthogonal}, or human driving situations \cite{Pegasus2019, Najm2007, DeGelder2020}.

\subsection{Validation of the SOTIF}\label{subsec:Val}

The Rulebooks' scenario-agnostic definition of driving behavior reduces the uncertainty around the nominal driving behavior in unforeseen scenarios, because the AV prioritizes the higher rules in the hierarchy in any scenario it encounters. This should result in better performance in unknown scenarios compared to an approach based on scenario-specific behavior specification, thus increasing the chances of success during statistical validation.  

Additionally, Rulebooks can help assess and improve the statistical performance of the AV when exposed to unknown scenarios (i.e., during statistically representative public road and simulation testing). Breaking down behavior into atomic elements (rules) makes it possible to relate poor performance to violation of individual rules. If an AV implementation proves unable to follow a Rulebook specification in many different situations, developers can identify the most difficult rules to follow. This can help tag scenarios encountered during validation testing with a level of criticality \cite{Junietz2018} based on the number and priority of violated rules. Tagging critical scenarios can guide on-road or virtual driving campaigns towards a more rapid reduction of the unknown unsafe scenarios. One can view this strategy as equivalent to a Markov process, obtained through simulation, used in adaptive stress testing in \cite{Corso2019}.

As discussed, changing the AV during the final stages of validation may invalidate the statistical inferences from this testing. Therefore, the latter advantage applies only during earlier stages of validation of the AV's SOTIF in unknown scenarios. However, the assessment of driving performance according to a Rulebook during the iterative testing can serve as a leading measure to provide confidence in the AV's SOTIF prior to the final statistical validation \cite{Fraade-blanar2018}. During the final statistical validation stage, a Rulebooks-based assessment of driving behavior can help detect any insufficiencies early on. This would not only avoid the many miles of public road driving it would take to uncover these insufficiencies through rare events like safety-critical takeovers, but also prevent the risk associated with these events.

The following section illustrates the concepts presented in this section by showing the application of a simplified Rulebook for the SOTIF process. 

\section{Example of use} \label{sec:Ex}

%\todo[inline]{the process for specification and validation of a sound rule set (before testing the driving function against it) does not become clear in the paper.}

\subsection{Functional Specification}
\begin{figure}[!h]
	\centering
	\includegraphics[scale=0.4]{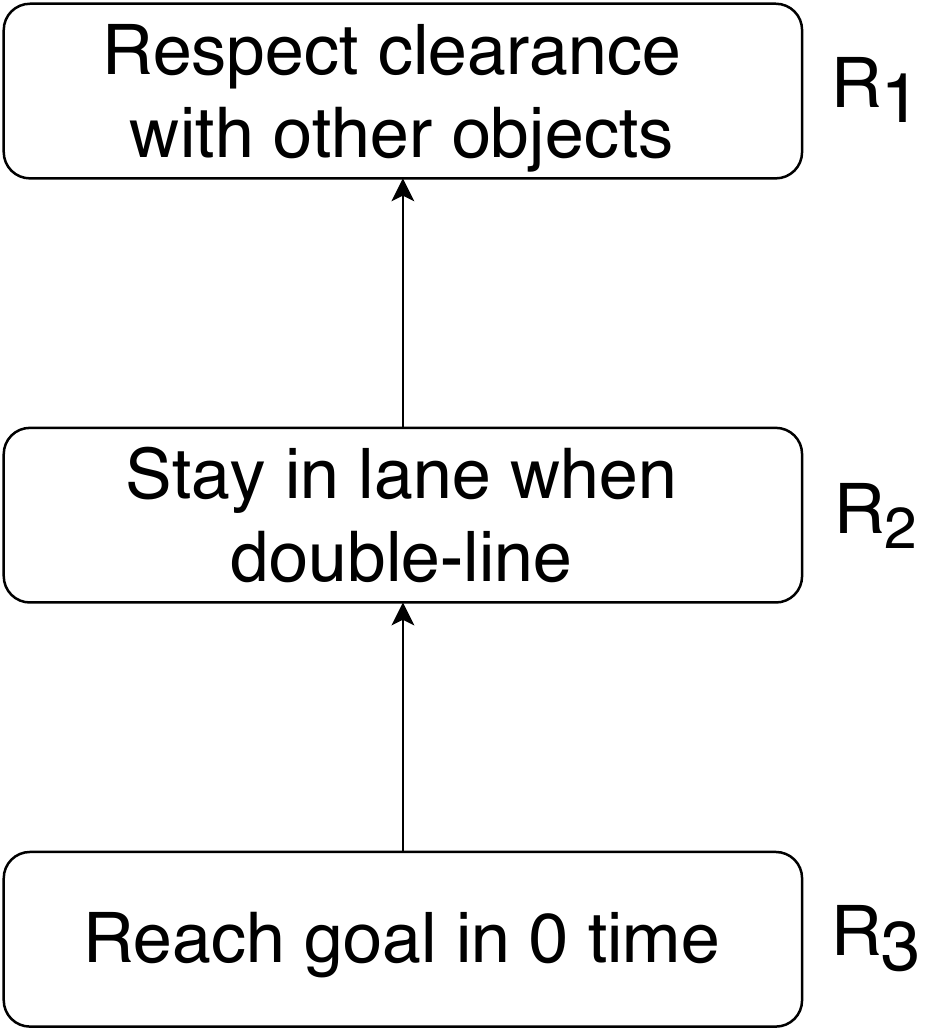}
	\caption{Example of a Rulebook, which is used to evaluate trajectories in this example. Developers might require the Rulebook to be a total order, but it should still align with the general pre-order presented in Figure \ref{fig:RBExample}.}
	\label{fig:MiniRB}
\end{figure}

To define a very simple, hypothetical driving functionality, we consider a minimal ODD that consists only of bidirectional roads with one lane in each direction. To specify the behavior, we select three rules with the following decreasing order of priority (Figure \ref{fig:MiniRB}):
\begin{itemize}
\item $R_1$: \textit{Respect clearance with other objects}. We give this rule the highest priority because breaking it directly relates to the probability of collision with other objects.
\item $R_2$: \textit{Stay in lane}. The AV should have a predictable behavior and stay in lane as much as possible, which also decreases the risk of collision with oncoming traffic. 
\item $R_3$: \textit{Reach goal in 0 time}. This rule is impossible to satisfy but is necessary to express a preference among trajectories that satisfy the other two rules (e.g., a preference for a trajectory that reaches the goal without delay versus a zero-speed trajectory). 
\end{itemize}

Rulebooks associate a violation metric with each of these rules. For $R_1$, one might consider the fraction of time steps during which the clearance with any object falls below the desired minimum clearance. Further refinements can compute an average over of the amount with which the AV falls below the desired minimum clearance, \cite{Mehdipour2019} and can make the desired minimum clearance a function of the type of object or the speed differential between the AV and the object. Similarly, the violation metric for $R_2$ can be the fraction of time steps during which we are outside of our lane, or a refinement thereof. $R_3$ may be the number of time steps required to reach the goal. The formulation of $R_3$ means that the AV is bound to violate the rule; this does not mean that the trajectory is inherently bad, it merely helps make the violation metric clear. A vaguer rule, such as "Reach goal in minimum time" would prevent quantifying the violation metric because the minimum time would depend on the scenario, whereas our formulation is scenario-agnostic.
With these three rules, we have fully specified the desired driving behavior of the hypothetical AV in this minimal ODD.

Defining the rule formulations, violation metrics, and priority structure is an iterative exercise. Before implementing the Rulebook on the AV or incorporating it in the systems engineering and SOTIF process, the AV developer may want to assess whether the Rulebook indeed specifies the desired driving behavior. The AV developer can do this by evaluating for a large number of trajectories in many different scenarios whether the Rulebook indeed results in rankings of trajectories that achieve the high-level requirements of the AV (e.g., safety, lawfulness, predictability, comfort, efficiency).

\subsection{Verification of the SOTIF}

\begin{figure}[!ht]
	\begin{subfigure}{.23\textwidth}
     %\subfloat[Scenario 1\label{fig:wide}]{%
       \includegraphics[height=6cm, width=\linewidth]{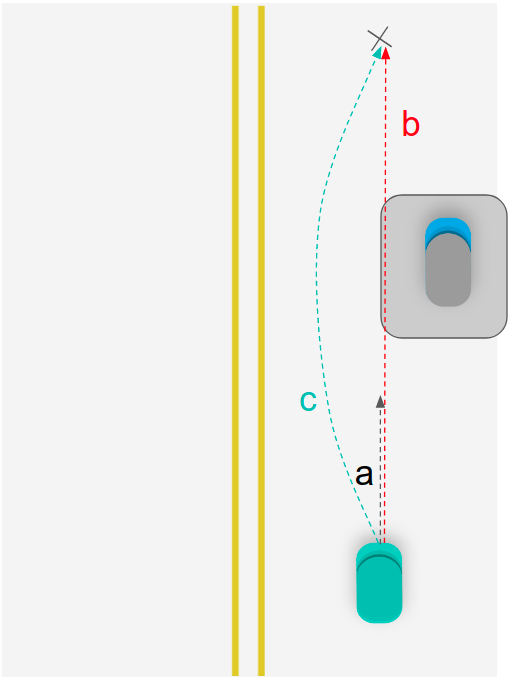}
       \caption{Scenario 1}
       \label{fig:wide}
     %}
     \end{subfigure}%
     \hfill
     \begin{subfigure}{.17\textwidth}
     %\subfloat[Scenario 2\label{fig:narrow}]{%
       \includegraphics[height=6cm, width=\linewidth]{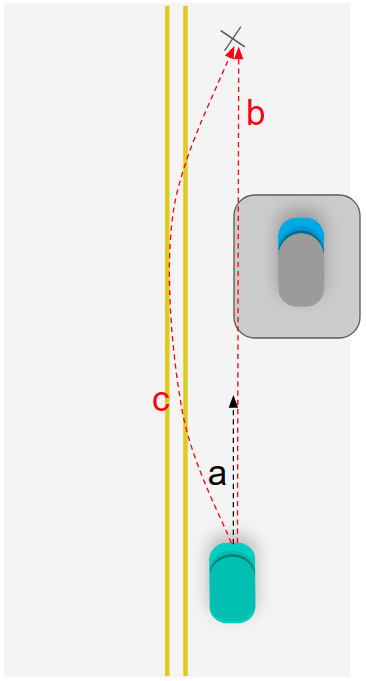}
       \caption{Scenario 2}
       \label{fig:narrow}
     %}
     \end{subfigure}
     \caption{Two similar driving scenarios, showing the ego vehicle in green, and a parked car in our lane in grey and clue. The grey box around the parked car shows the zone we need to avoid to maintain clearance. A traffic law prohibits crossing of the lane boundary. In both scenarios, the black cross at the top of the lane is the goal. The only difference between the two scenarios is the lane width.}
     \label{fig:parkedCarScenario}
   \end{figure}

Verification of the SOTIF considers known scenarios. For purposes of illustration, we here consider a very limited sample. Figure \ref{fig:parkedCarScenario} shows two driving scenarios. Both scenarios involve  a parked car in our lane and assume that the AV's goal lies just beyond the parked car. We further assume that there exists a traffic law prohibiting lane boundary  crossings (double line in the United States), which traces to $R_2$. In Scenario 1 (Figure \ref{fig:wide}), our lane is wide enough to pass the parked car while respecting the desired clearance with it. In contrast, in scenario 2 (Figure \ref{fig:narrow}), the lane is too narrow. To pass the parked car, the AV would have to either cross the lane separation or not satisfy the clearance rule.

We evaluate three trajectories in both scenarios: in trajectory (a), the ego vehicle goes straight, decelerates and stops before reaching the parked car on its right side. In trajectory (b), the ego vehicle continues to drive straight, not respecting the clearance rule for several time steps in both scenarios. In trajectory (c), the ego vehicle biases to its left, respecting the clearance rule with the parked vehicle in both scenarios. However, this behavior results in crossing the lane boundary in Scenario 2.

As described in \cite{Censi2019}, Rulebooks use lexicographical comparisons to determine which trajectory is best by going down the Rulebooks priority structure. In Scenario 1, (b) violates $R_1$ while (a) and (c) do not, so (b) is the least preferred trajectory. (a) and (c) both respect $R_2$, but (a)  violates $R_3$ by a larger amount than (c). The ranking of trajectories in Scenario 1 is therefore $c>a>b$. Note that violating a rule by itself is not a reason for rejection, as all of the trajectories violate $R_3$.

Using the same process for Scenario 2; (b) still violates $R_1$ while (a) and (c) do not, eliminating (b). (c) now violates $R_2$ while (a) does not. Thus, the ranking of trajectories in Scenario 2 is $a>c>b$. The best trajectory (a) stops indefinitely, which is a logical consequence of the behavior specified in Figure \ref{fig:MiniRB}. If this was in fact not the behavior that we desired, then we would need to refine the rule priority structure or the violation metrics. For example, the desired minimum clearance can be a function of the ego-vehicle speed instead of being a constant. This could allow passing the parked car at a low speed. This step corresponds to Clause 8 in ISO/PAS 21448, in which one modifies the functional definition to reduce the SOTIF risk, or here to increase satisfaction of the stakeholder need to get to the goal without sacrificing the SOTIF.

This simplified example demonstrates the power of having a general framework. With only three rules, the nominal behavior for the ego vehicle would be lawful in two scenarios which might seem similar from a geometric perspective, but in reality prompt two different driving behaviors. We can construct a ranking of trajectories without having to specify a large number of conditions by hand, such as "\textit{if the lane is not wide enough, then stop}". Instead, the desired passing or stopping behavior is a direct result of the intersection of hierarchical rules.

An actual AV implementation of the Rulebook (i.e., through direct implementation in the planner and/or system requirements derived from the behavior specification) might not exactly follow the highest-ranking trajectory because of implementation imperfections. However, the Rulebooks framework offers a method to score simulated or on-road vehicle behavior. Since the higher rules are more critical, a passing criterion for an AV trajectory might be that it does not violate either $R_1$ or $R_2$. If the AV's trajectory in a verification test is similar to (b) in both scenarios, then the AV would fail the behavior assessment. This may lead to appropriate changes of the architecture or the system components to improve performance in future iterations.

\subsection{Validation of the SOTIF}

The performance of the vehicle with regards to driving behavior can be an indicator for the probability of exhibiting the correct behavior in unforeseen scenarios. Now that the behavior is explainable and traceable to atomic elements, we can identify unsafe scenarios. 
%As describes in section \ref{subsec:Val}, the method to generate new unsafe scenario is the following:
%\begin{enumerate}
%\item Classify scenarios in which the vehicle's behavior is unsatisfactory as unsafe
%\item Determine the rule, or rules, involved in the unsafe scenarios
%\item Generate scenarios involving those specific rules.
%\end{enumerate}
%%\todo[inline]{you said extrapolate instead of identify, but I'm not sure what that meant}

For example, if the ego vehicle exhibits higher violation scores in the simulation tests involving the clearance rule, then engineers know that this specific rule is more difficult to adhere to for the AV than other rules. We can therefore assume that future scenarios involving violation of this rule are more difficult for the AV, and we can focus on them to more rapidly assess and increase the safety of the AV. We can also trace the difficulty with clearance back to a specific component of the system, for example to performance limitations in the mapping and localization subsystem.

As mentioned in subsection \ref{subsec:Val}, one can also use the Rulebooks framework to rapidly assess the SOTIF.  For example, frequent violations of $R_1$ may manifest much sooner than actual safety-critical takeover incidents in which the AV comes dangerously close to other objects. Conversely, observing consistent satisfaction of $R_1$ and $R_2$ in previously unknown scenarios that arise during public road testing (or during exposure of the AV to new scenarios in simulation) provides confidence that the AV's driving behavior is sufficient to proceed to a statistical safety validation mileage accumulation effort.

\section{Discussion and Conclusion}\label{sec:Concl}

This paper shows how AV developers may apply the Rulebooks framework as part of their SOTIF process. Rulebooks specify driving behavior through the precise definition of rules and a priority structure among rules. The violation metrics on trajectories that a Rulebook provides can guide the verification process, and can help identify specific unsafe scenarios, to speed up vehicle driving behavior validation.

The Rulebooks approach presented in this paper is not only scenario-agnostic, but also implementation-free. Indeed, we hope that specifying priorities among driving rules can help prompt a conversation about the principles that should guide AVs, no matter which vehicle platform or algorithmic choices specific companies make. Although a Rulebooks behavior specification is implementation-free, one can always incorporate the limitations associated with a specific design into a practical version of a Rulebook. For example, a practical version would not allow the vehicle to drive as fast if it does not have good recognition or prediction capabilities. Furthermore, the Rulebooks approach does not mandate any specific algorithm. Any solution for object detection, behavior prediction, or motion planning is possible provided that the resulting behavior respects the rules in their hierarchy.

The main limitations of this method are the time and resources required to come to an acceptable rule set and priority structure. Testing on the road and simulations might prompt a reconsideration of the rules or their hierarchy, not unlike a systems requirement writing exercise. However, working on these issues at the development stage will likely save time by building in good behaviors from the design stages of the AV.

Although the sole use of this method is not sufficient to demonstrate safety, the authors believe it offers an important, concrete step towards fulfilling the recommendations of ISO/PAS 21448. This method does not require separate behavior definitions for each scenario that may occur in an ODD. The framework further provides transparency, traceability, and explainability of an AV's driving behavior. 

Future work may explore how to best capture the interest of all the stakeholders of autonomous driving, including other road users, policy makers, local communities and the AV industry. Rulebooks take the burden of AV behavior definition out of the development teams and open it up for a broader discussion of how society would like AVs to behave. 

\addtolength{\textheight}{-14.75cm}

\bibliographystyle{BibFiles/IEEEtranUrldate.bst}
\bibliography{BibFiles/IEEEabrv,IV2020}

\end{document}